\documentclass{article}
\usepackage{ijcai19}

\usepackage{times}
\usepackage{soul}
\usepackage{url}
\usepackage[hidelinks]{hyperref}
\usepackage[utf8]{inputenc}
\usepackage[small]{caption}
\usepackage{graphicx}
\usepackage{amsmath}
\usepackage{booktabs}
\usepackage{algorithm}
\usepackage{algorithmic}
\usepackage{amsmath, amssymb, amsthm, amsfonts}
\usepackage{multirow}
\usepackage{subcaption}
\usepackage{xcolor}

\newtheorem{definition}{Definition}
\newtheorem{proposition}{Proposition}

\title{Federated Generative Privacy}

\author{
Aleksei Triastcyn
\and
Boi Faltings
\affiliations
Artificial Intelligence Lab \\ Ecole Polytechnique F\'ed\'erale de Lausanne \\ Lausanne, Switzerland
\emails
\{aleksei.triastcyn, boi.faltings\}@epfl.ch,
}

\begin{document}

\maketitle

\begin{abstract}
In this paper, we propose \texttt{FedGP}, a framework for privacy-preserving data release in the federated learning setting. We use generative adversarial networks, generator components of which are trained by \texttt{FedAvg} algorithm, to draw privacy-preserving artificial data samples and empirically assess the risk of information disclosure. Our experiments show that \texttt{FedGP} is able to generate labelled data of high quality to successfully train and validate supervised models. Finally, we demonstrate that our approach significantly reduces vulnerability of such models to model inversion attacks.
\end{abstract}

\section{Introduction}
\label{sec:introduction}
The rise of data analytics and machine learning (ML) presents countless opportunities for companies, governments and individuals to benefit from the accumulated data. At the same time, their ability to capture fine levels of detail potentially compromises privacy of data providers. Recent research~\cite{fredrikson2015model,shokri2017membership,hitaj2017deep} suggests that even in a black-box setting it is possible to argue about the presence of individual examples in the training set or recover certain features of these examples.

Among methods that tackle privacy issues of machine learning is the recent concept of \emph{federated learning} (FL)~\cite{mcmahan2016communication}. In the FL setting, a central entity (\emph{server}) wants to train a model on user data without actually copying these data from user devices. Instead, users (\emph{clients}) update models locally, and the \emph{server} aggregates these models. One popular approach is the federated averaging, \texttt{FedAvg}~\cite{mcmahan2016communication}, where \emph{clients} do local on-device gradient descent using their data, then send these updates to the \emph{server} where they get averaged. Privacy can further be enhanced by using secure multi-party computation (MPC)~\cite{yao1982protocols} to allow the server access only average updates of a big group of users and not individual ones.

Despite many advantages, federated learning does have a number of challenges. First, the result of FL is a single trained model (therefore, we will refer to it as a \emph{model release} method), which does not provide much flexibility in the future. For instance, it would significantly reduce possibilities for further aggregation from different sources, e.g. different hospitals trying to combine federated models trained on their patients data. Second, this solution requires data to be labelled at the source, which is not always possible, because user may be unqualified to label their data or unwilling to do so. A good example is again a medical application where users are unqualified to diagnose themselves but at the same time would want to keep their condition private. Third, it does not provide provable privacy guarantees, and there is no reason to believe that the aforementioned attacks do not work against it. Some papers propose to augment FL with differential privacy (DP) to alleviate this issue~\cite{mcmahan2017learning,geyer2017differentially}. While these approaches perform well in ML tasks and provide theoretical privacy guarantees, they are often restrictive (e.g. many DP methods for ML assume, implicitly or explicitly, access to public data of similar nature or abundant amounts of data, which is not always realistic).

\begin{figure}
	\centering
	\includegraphics[width=\linewidth]{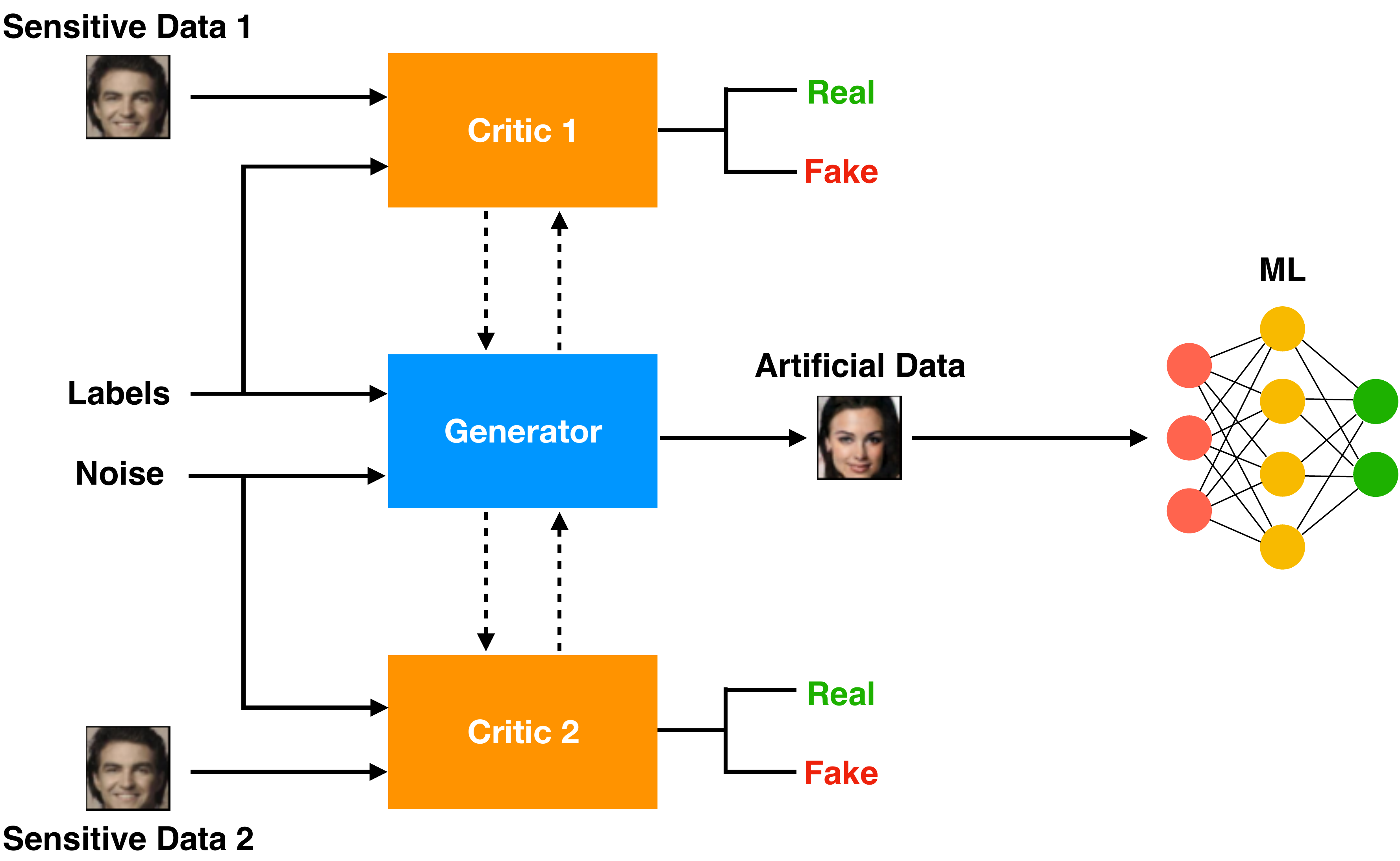}
    	\caption{Architecture of our solution for two clients. Sensitive data is used to train a GAN (local critic and federated generator) to produce a private artificial dataset, which can be used by any ML model.}
	\label{fig:architecture}
\end{figure}

In our work, we address these problems by proposing to combine the strengths of federated learning and recent advancements in generative models to perform privacy-preserving \emph{data release}, which has many immediate advantages. First, the released data could be used to train any ML model (we refer to it as the \emph{downstream task} or the \emph{downstream model}) without additional assumptions. Second, data from different sources could be easily pooled, providing possibilities for hierarchical aggregation and building stronger models. Third, labelling and verification can be done later down the pipeline, relieving some trust and expertise requirements on users. Fourth, released data could be traded on data markets\footnote{\url{https://www.datamakespossible.com/value-of-data-2018/dawn-of-data-marketplace}}, where anonymisation and protection of sensitive information is one of the biggest obstacles. Finally, data publishing would facilitate transparency and reproducibility of research studies.

The main idea of our approach, named \texttt{FedGP}, for \emph{federated generative privacy}, is to train generative adversarial networks (GANs)~\cite{goodfellow2014generative} on clients to produce artificial data that can replace clients real data. Since some clients may have insufficient data to train a GAN locally, we instead train a federated GAN model. First of all, user data still remain on their devices. Second, the federated GAN will produce samples from the common cross-user distribution and not from a specific single user, which adds to overall privacy. Third, it allows releasing entire datasets, thereby possessing all the benefits of private \emph{data release} as opposed to \emph{model release}. Figure~\ref{fig:architecture} depicts the schematics of our approach for two clients.

To estimate potential privacy risks, we use our \emph{post hoc} privacy analysis framework~\cite{triastcyn2019generating} designed specifically for private data release using GANs.

Our contributions in this paper are the following:
\begin{itemize}
\item on the one hand, we extend our approach for private data release to the federated setting, broadening its applicability and enhancing privacy;
\item on the other hand, we modify the federated learning protocol to allow a range of benefits mentioned above;
\item we demonstrate that downstream models trained on artificial data achieve high learning performance while maintaining good average-case privacy and being resilient to model inversion attacks.
\end{itemize}

The rest of the paper is structured as follows. In Section~\ref{sec:related_work}, we give an overview of related work. Section~\ref{sec:preliminaries} contains some preliminaries. In Section~\ref{sec:approach}, we describe our approach and privacy estimation framework. Experimental results are presented in Section~\ref{sec:evaluation}, and Section~\ref{sec:conclusion} concludes the paper.

\section{Related Work}
\label{sec:related_work}

In recent years, as machine learning applications become a commonplace, a body of work on security of these methods grows at a rapid pace. Several important vulnerabilities and corresponding attacks on ML models have been discovered, raising the need of devising suitable defences. Among the attacks that compromise privacy of training data, model inversion~\cite{fredrikson2015model} and membership inference~\cite{shokri2017membership} received high attention.

Model inversion~\cite{fredrikson2015model} is based on observing the output probabilities of the target model for a given class and performing gradient descent on an input reconstruction. Membership inference~\cite{shokri2017membership} assumes an attacker with access to similar data, which is used to train a "shadow" model, mimicking the target, and an attack model. The latter predicts if a certain example has already been seen during training based on its output probabilities. Note that both attacks can be performed in a black-box setting, without access to the model internal parameters.

To protect privacy while still benefiting from the use of statistics and ML, many techniques have been developed over the years, including $k$-anonymity~\cite{sweeney2002}, $l$-diversity~\cite{machanavajjhala2007}, $t$-closeness~\cite{li2007t}, and differential privacy (DP)~\cite{dwork2006}.

Most of the ML-specific literature in the area concentrates on the task of privacy-preserving model release. One take on the problem is to distribute training and use disjoint datasets. For example,  \cite{shokri2015privacy} propose to train a model in a distributed manner by communicating sanitised updates from participants to a central authority. Such a method, however, yields high privacy losses~\cite{abadi2016deep,papernot2016semi}. An alternative technique suggested by \cite{papernot2016semi}, also uses disjoint training sets and builds an ensemble of independently trained teacher models to transfer knowledge to a student model by labelling public data. This result has been extended in \cite{papernot2018scalable} to achieve state-of-the-art image classification results in a private setting (with single-digit DP bounds).
A different approach is taken by \cite{abadi2016deep}. They suggest using differentially private stochastic gradient descent (DP-SGD) to train deep learning models in a private manner. This approach achieves high accuracy while maintaining low DP bounds, but may also require pre-training on public data.

A more recent line of research focuses on private data release and providing privacy via generating synthetic data~\cite{bindschaedler2017plausible,huang2017context,beaulieu2017privacy}. In this scenario, DP is hard to guarantee, and thus, such models either relax the DP requirements or remain limited to simple data. In \cite{bindschaedler2017plausible}, authors use a graphical probabilistic model to learn an underlying data distribution and transform real data points (seeds) into synthetic data points, which are then filtered by a privacy test based on a \emph{plausible deniability} criterion. This procedure would be rather expensive for complex data, such as images. \citeauthor{fioretto2019privacy}~\shortcite{fioretto2019privacy} employ dicision trees for a hybrid model/data release solution and guarantee stronger $\varepsilon$-differential privacy, but like the previous approach, it would be difficult to adapt to more complex data. Alternatively, \citeauthor{huang2017context}~\shortcite{huang2017context} introduce the notion of \emph{generative adversarial privacy} and use GANs to obfuscate real data points w.r.t. pre-defined private attributes, enabling privacy for more realistic datasets. Finally, a natural approach to try is training GANs using DP-SGD~\cite{beaulieu2017privacy,xie2018differentially,zhang2018differentially}. However, it proved extremely difficult to stabilise training with the necessary amount of noise, which scales as $\sqrt{m}$ w.r.t. the number of model parameters $m$. It makes these methods inapplicable to more complex datasets without resorting to unrealistic (at least for some areas) assumptions, like access to public data from the same distribution.

On the other end of spectrum, \citeauthor{mcmahan2016communication}~\shortcite{mcmahan2016communication} proposed federated learning as one possible solution to privacy issues (among other problems, such as scalability and communication costs). In this setting, privacy is enforced by keeping data on user devices and only submitting model updates to the server. It can be augmented by MPC~\cite{bonawitz2017practical} to prevent the server from accessing individual updates and by DP~\cite{mcmahan2017learning,geyer2017differentially} to provide rigorous theoretical guarantees.

\section{Preliminaries}
\label{sec:preliminaries}

This section provides necessary definitions and background. Let us commence with approximate differential privacy.

\begin{definition}
A randomised function (mechanism) $\mathcal{M}: \mathcal{D} \rightarrow \mathcal{R}$ with domain $\mathcal{D}$ and range $\mathcal{R}$ satisfies $(\varepsilon, \delta)$-differential privacy if for any two adjacent inputs $d, d' \in \mathcal{D}$ and for any outcome $o \in \mathcal{R}$ the following holds:
\begin{align}
	\Pr\left[\mathcal{M}(d)=o\right] \leq e^\varepsilon \Pr\left[\mathcal{M}(d') = o\right] + \delta.
\end{align}
\end{definition}

\begin{definition}
Privacy loss of a randomised mechanism $\mathcal{M}: \mathcal{D} \rightarrow \mathcal{R}$ for inputs $d, d' \in \mathcal{D}$ and outcome $o \in \mathcal{R}$ takes the following form:
\begin{align}
	L_{(\mathcal{M}(d) \| \mathcal{M}(d'))} = \log\frac{\Pr\left[\mathcal{M}(d) = o \right]}{\Pr\left[\mathcal{M}(d') = o \right]}.
\end{align}
\end{definition}

\begin{definition}
The Gaussian noise mechanism achieving $(\varepsilon, \delta)$-DP, for a function $f: \mathcal{D} \rightarrow \mathbb{R}^m$, is defined as
\begin{align}
	\mathcal{M}(d) = f(d) + \mathcal{N}(0, \sigma^2),
\end{align}
where $\sigma > C \sqrt{2\log\frac{1.25}{\delta}} / \varepsilon$ and $C$ is the L2-sensitivity of $f$.
\end{definition}

For more details on differential privacy and the Gaussian mechanism, we refer the reader to~\cite{dwork2014algorithmic}.

In our privacy estimation framework, we also use some classical notions from probability and information theory.

\begin{definition}
The Kullback–Leibler (KL) divergence between two continuous probability distributions $P$ and $Q$ with corresponding densities $p$, $q$ is given by:
\begin{align}
	D_{KL}(P \| Q) = \int_{-\infty}^{+\infty} p(x) \log\frac{p(x)}{q(x)} dx.
\end{align}
\end{definition}

Note that KL divergence between the distributions of $\mathcal{M}(d)$ and $\mathcal{M}(d')$ is nothing but the expectation of the privacy loss random variable $\mathbb{E}[L_{(\mathcal{M}(d) \| \mathcal{M}(d'))}]$.

Finally, we use the Bayesian perspective on estimating mean from the data to get sharper bounds on expected privacy loss compared to the original work~\cite{triastcyn2019generating}. More specifically, we use the following proposition.
\begin{proposition}
\label{prop:bayesian}
Let $[l_1, l_2, \ldots, l_n]$ be a random vector drawn from the distribution $p(L)$ with the same mean and variance, and let $\overline{L}$ and $S$ be the sample mean and the sample standard deviation of the random variable $L$. Then,
\begin{align}
	\Pr\left(\mathbb{E}[L] > \overline{L} + \frac{F_{n-1}^{-1}(1 - \gamma)}{\sqrt{n-1}} S \right) \leq \gamma,
\end{align}
where $F_{n-1}^{-1}(1 - \gamma)$ is the inverse CDF of the Student's $t$-distribution with $n-1$ degrees of freedom at $1 - \gamma$.
\end{proposition}
The proof of this proposition can be obtained by using the maximum entropy principle with a flat (uninformative) prior to get the marginal distribution of the sample mean $\overline{L}$, and observing that the random variable $\frac{\mathbb{E}[L] - \overline{L}}{S/\sqrt{n-1}}$ follows the Student's $t$-distribution with $n - 1$ degrees of freedom~\cite{oliphant2006bayesian}.

\section{Federated Generative Privacy}
\label{sec:approach}
In this section, we describe our algorithm, what privacy it can provide and how to evaluate it, and discuss current limitations.

\subsection{Method Description}
In order to keep participants data private while still maintaining flexibility in downstream tasks, our algorithm produces a federated generative model. This model can output artificial data, not belonging to any real user in particular, but coming from the common cross-user data distribution.

Let $\{u_1, u_2, \ldots, u_n\}$ be a set of \emph{clients} holding private datasets $\{d_1, d_2, \ldots, d_n\}$. Before starting the training protocol, the \emph{server} is providing each \emph{client} with generator $G_i^0$ and critic $C_i^0$ models, and \emph{clients} initialise their models randomly. Like in a normal FL setting, the training process afterwords consists of communication rounds. In each round $t$, \emph{clients} update their respective models performing one or more passes through their data and submit generator updates $\triangle G_i^t$ to the \emph{server} through MPC while keeping $C_i^t$ private. In the beginning of the next round, the \emph{server} provides an updated common generator $G^t$ to all \emph{clients}.

This approach has a number of important advantages:
\begin{itemize}
\item Data do not physically leave user devices.
\item Only generators (that do not come directly into contact with data) are shared, and critics remain private.
\item Using artificial data in downstream tasks adds another layer of protection and limits the information leakage to artificial samples. This is esprecially useful given that ML models can be attacked to extract training data~\cite{fredrikson2015model}, sometimes even when protected by DP~\cite{hitaj2017deep}.
\end{itemize}

What remains to assess is how much information would an attacker gain about original data. We do so by employing a notion introduced in an earlier work~\cite{triastcyn2019generating} that we name \emph{Differential Average-Case Privacy}.

It is important to clarify why we do not use the standard DP to provide stronger theoretical guarantees: we found it extremely difficult to train GANs with the amount of noise required for meaningful DP guarantees. Despite a number of attempts~\cite{beaulieu2017privacy,xie2018differentially,zhang2018differentially}, we are not aware of any technically sound solution that would generalise beyond very simple datasets.

\subsection{Differential Average-Case Privacy}
\label{sec:dap}
Our framework builds upon ideas of \emph{empirical DP} (EDP)~\cite{abowd2013differential,schneider2015new} and \emph{on-average KL privacy}~\cite{wang2016average}. The first can be viewed as a measure of sensitivity on posterior distributions of outcomes~\cite{charest2017meaning} (in our case, generated data distributions), while the second relaxes DP notion to the case of an average user.

More specifically, we say the mechanism $\mathcal{M}$ is $(\mu, \gamma)$-DAP if for two neighbouring datasets $D, D'$, where data come from an observed distribution, it holds that
\begin{align}
\label{eq:dap}
	\Pr(\mathbb{E}[|L_{(\mathcal{M}(D) \| \mathcal{M}(D'))}|] > \mu) \leq \gamma.
\end{align}

For the sake of example, let each data point in $D, D'$ represent a single user. Then, $(0.01, 0.001)$-DAP could be interpreted as follows: with probability $0.999$, a typical user submitting their data will change outcome probabilities of the private algorithm on average by $1\%$\footnote{Because $e^{0.01} \approx 1.01.$}.

\subsection{Generative Differential Average-Case Privacy}
\label{sec:gdap}
In the case of generative models, and in particular GANs, we don't have access to exact posterior distributions, a straightforward EDP procedure in our scenario would be the following:
\emph{(1)} train GAN on the original dataset $D$; 
\emph{(2)} remove a random sample from $D$; 
\emph{(3)} re-train GAN on the updated set; 
\emph{(4)} estimate probabilities of all outcomes and the maximum privacy loss value; 
\emph{(5)} repeat \emph{(1)--(4)} sufficiently many times to approximate $\varepsilon$, $\delta$.

If the generative model is simple, this procedure can be used without modification. Otherwise, for models like GANs, it becomes prohibitively expensive due to repetitive re-training (steps \emph{(1)--(3)}). Another obstacle is estimating the maximum privacy loss value (step \emph{(4)}). To overcome these two issues, we propose the following.

First, to avoid re-training, we imitate the removal of examples directly on the generated set $\widetilde{D}$. We define a similarity metric $sim(x, y)$ between two data points $x$ and $y$ that reflects important characteristics of data (see Section~\ref{sec:evaluation} for details). For every randomly selected real example $i$, we remove $k$ nearest artificial neighbours to simulate absence of this example in the training set and obtain $\widetilde{D}^{-i}$. Our intuition behind this operation is the following. Removing a real example would result in a lower probability density in the corresponding region of space. If this change is picked up by a GAN, which we assume is properly trained (e.g. there is no mode collapse), the density of this region in the generated examples space should also decrease. The number of neighbours $k$ is defined by the ratio of artificial and real examples, to keep density normalised.

Second, we relax the worst-case privacy loss bound in step \emph{(4)} by the expected-case bound, in the same manner as on-average KL privacy. This relaxation allows us to use a high-dimensional KL divergence estimator~\cite{perez2008kullback} to obtain the expected privacy loss for every pair of adjacent datasets ($\widetilde{D}$ and $\widetilde{D}^{-i}$). There are two major advantages of this estimator: it converges almost surely to the true value of KL divergence; and it does not require intermediate density estimates to converge to the true probability measures. Also since this estimator uses nearest neighbours to approximate KL divergence, our heuristic described above is naturally linked to the estimation method.

Finally, having obtained sufficiently many sample pairs $(\widetilde{D}, \widetilde{D}^{-i})$, we use Proposition~\ref{prop:bayesian} to determine DAP parameters $\mu$ and $\gamma$. This is an improvement over original DAP, because this way we can get much sharper bounds on expected privacy loss.

\subsection{Limitations}
\label{sec:limitations}
Our approach has a number of limitations that should be taken into consideration. 

First of all, existing limitations of GANs (or generative models in general), such as training instability or mode collapse, will apply to this method. Hence, at the current state of the field, our approach may be difficult to adapt to inputs other than image data. Yet, there is still a number of privacy-sensitive applications, e.g. medical imaging or facial analysis, that could benefit from our technique. And as generative methods progress, new uses will be possible.

Second, since critics remain private and do not leave user devices their performance can be hampered by a small number of training examples. Nevertheless, we observe that even in the setting where some users have smaller datasets overall discriminative ability of all critics is sufficient to train good generators.

Lastly, our empirical privacy guarantee is not as strong as the traditional DP and has certain limitations \cite{charest2017meaning}. However, due to the lack of DP-achieving training methods for GANs it is still beneficial to have an idea about expected privacy loss rather than not having any guarantee.

\section{Evaluation}
\label{sec:evaluation}

\begin{table}
	\caption{Accuracy of student models trained on artificial samples of FedGP compared to non-private centralised baseline and CentGP. In parenthesis we specify the average number of data points per client. }
	\label{tab:accuracy}
	\centering
	\begin{tabular}{ l l c c c }
		\toprule
		{\bf Setting} 												& {\bf Dataset}		& {\bf Baseline}		& {\bf CentGP} 	& {\bf FedGP} \\
		\midrule
		\multirow{3}{*}{i.i.d.}									& MNIST (500)		& $98.10\%$ 		& $97.35\%$ 		& $79.45\%$ \\
																		& MNIST (1000)	& $98.55\%$ 		& $97.39\%$ 		& $93.38\%$ \\
																		& MNIST (2000)	& $98.92\%$ 		& $97.41\%$ 		& $96.23\%$ \\
		\midrule
		\multirow{3}{*}{\shortstack{non-\\i.i.d.}}	& MNIST (500)		& $97.31\%$ 		& 							& $83.26\%$ \\
																		& MNIST (1000)	& $98.78\%$ 		&	---					& $95.89\%$ \\
																		& MNIST (2000)	& $98.76\%$ 		&							& $96.88\%$ \\
		\bottomrule
	\end{tabular}
\end{table}

In this section, we describe the experimental setup and implementation, and evaluate our method on MNIST~\cite{lecun1998gradient} and CelebA~\cite{liu2015faceattributes} datasets.

\subsection{Experimental Setting}

We evaluate two major aspects of our method. First, we show that training ML models on data created by the common generator achieves high accuracy on MNIST (Section~\ref{sec:learning}). Second, we estimate expected privacy loss of the federated GAN and evaluate the effectiveness of artificial data against model inversion attacks on CelebA face attributes (Section~\ref{sec:privacy}).

Learning performance experiments are set up as follows:
\begin{enumerate}
\item Train the federated generative model (\emph{teacher}) on the original data distributed across a number of users.
\item Generate an artificial dataset by the obtained model and use it to train ML models (\emph{students}).
\item Evaluate students on a held-out test set. 
\end{enumerate}

We choose two commonly used image datasets, MNIST and CelebA. MNIST is a handwritten digit recognition dataset consisting of 60000 training examples and 10000 test examples, each example is a 28x28 size greyscale image. CelebA is a facial attributes dataset with 202599 images, each of which we crop to 128x128 and then downscale to 48x48.

In our experiments, we use Python and Pytorch framework.\footnote{\url{http://pytorch.org}} For implementation details of GANs and privacy evaluation, please refer to \cite{triastcyn2019generating}. To train the federated generator we use FedAvg algorithm~\cite{mcmahan2016communication}. As a \emph{sim} function introduced in Section~\ref{sec:gdap} we use the distance between InceptionV3~\cite{szegedy2016rethinking} feature vectors.

\begin{table}
	\caption{Average-case privacy parameters: expected privacy loss bound $\mu$ and probability $\gamma$ of exceeding it.}
	\label{tab:privacy}
	\centering
	\begin{tabular}{l l c c c}
		\toprule
		{\bf Setting} 					& {\bf Dataset}			&~	& $\mu$  		& $\gamma$ \\
		\midrule
		\multirow{2}{*}{i.i.d.} 	& MNIST (500) 			&~	& $0.0117$ 		& \multirow{4}{*}{$10^{-15}$} \\
											& MNIST (1000) 		&~	& $0.0069$ 		&  \\
											& MNIST (2000) 		&~	& $0.0021$ 		&  \\
											& CelebA					&~	& $0.0009$ 		&  \\
		\midrule
		non-i.i.d.						& MNIST (500) 			&~	& $0.0090$ 		& \multirow{3}{*}{$10^{-15}$}  \\
											& MNIST (1000) 		&~	& $0.0044$ 		& \\
											& MNIST (2000) 		&~	& $0.0020$ 		& \\
		\bottomrule
	\end{tabular}
\end{table}

\iffalse
\begin{table}
	\caption{Empirical privacy parameters: expected privacy loss bound $\mu$ and probability $\gamma$ of exceeding it.}
	\label{tab:privacy}
	\centering
	\begin{tabular}{l l c c c}
		\toprule
		{\bf Setting} 					& {\bf Dataset}		& $\mu_c$		& $\mu_b$  		& $\gamma$ \\
		\midrule
		\multirow{2}{*}{i.i.d.} 	& MNIST (500) 		& $3.6858$ 		& $0.0117$ 		& \multirow{4}{*}{$10^{-5}$} \\
											& MNIST (1000) 	& $1.8856$ 		& $0.0069$ 		&  \\
											& MNIST (2000) 	& $0.6469$ 		& $0.0021$ 		&  \\
											& CelebA				& $0.2703$ 		& $0.0009$ 		&  \\
		\midrule
		non-i.i.d.						& MNIST (500) 		& $2.8393$		& $0.0090$ 		& \multirow{3}{*}{$10^{-5}$}  \\
											& MNIST (1000) 	& $1.3777$ 		& $0.0044$ 		& \\
											& MNIST (2000) 	& $0.6911$ 		& $0.0020$ 		& \\
		\bottomrule
	\end{tabular}
\end{table}
\fi

\subsection{Learning Performance}
\label{sec:learning}
First, we evaluate the generalisation ability of the student model trained on artificial data. More specifically, we train a student model on generated data and report test classification accuracy on a held-out real set. We compare learning performance with the baseline centralised model trained on original data, as well as the same model trained on artificial samples obtained from the centrally trained GAN (\texttt{CentGP}).

Since critics stay private and can only access data of a single user, the size of each individual dataset has significant effect. Therefore, in our experiment we vary sizes of user datasets and observe its influence on training. In each experiment, we specify an average number of points per user, while the actual number is drawn from the uniform distribution with this mean, with some clients getting as few as 100 data points.

We also study two settings: i.i.d. and non-i.i.d data.  In the first setting, distribution of classes for each client is identical to the overall distribution. In the second, every client gets samples of 2 random classes, imitating the situation when a single user observes only a part of overall data distribution.

Details of the experiment can be found in Table~\ref{tab:accuracy}. We observe that training on artificial data from the federated GAN allows to achieve $96.9\%$ accuracy on MNIST with the baseline of $98.8\%$. We can also see how accuracy grows with the average user dataset size. A less expected observation is that non-i.i.d. setting is actually beneficial for \texttt{FedGP}. A possible reason is that training critics with little data becomes easier when this data is less diverse (i.e. the number of different classes is smaller). Comparing to the centralised generative privacy model \texttt{CentGP}, we can also see that \texttt{FedGP} is more affected by sharding of data on user devices than by overall data size, suggesting that further research in training federated generative models is necessary.

\subsection{Privacy Analysis}
\label{sec:privacy}

Using the privacy estimation framework (see Sections~\ref{sec:dap} and~\ref{sec:gdap}), we fix the probability $\gamma$ of exceeding the expected privacy loss bound $\mu$ in all experiments to $10^{-15}$ and compute the corresponding $\mu$ for each dataset and two settings. Table~\ref{tab:privacy} summarises the bounds we obtain. As anticipated, the privacy guarantee improves with the growing number of data points, because the influence of each individual example diminishes. Moreover, the average privacy loss $\mu$, expectedly, is significantly smaller than the typical worst-case DP loss $\varepsilon$ in similar settings. To put it in perspective, the average change in outcome probabilities estimated by DAP is ${\sim}1\%$ even in more difficult settings, while the state-of-the-art DP method would place the worst-case change at hundreds or even thousands percent without giving much information about a typical case.

\begin{figure}
	\centering
	\includegraphics[width=\linewidth]{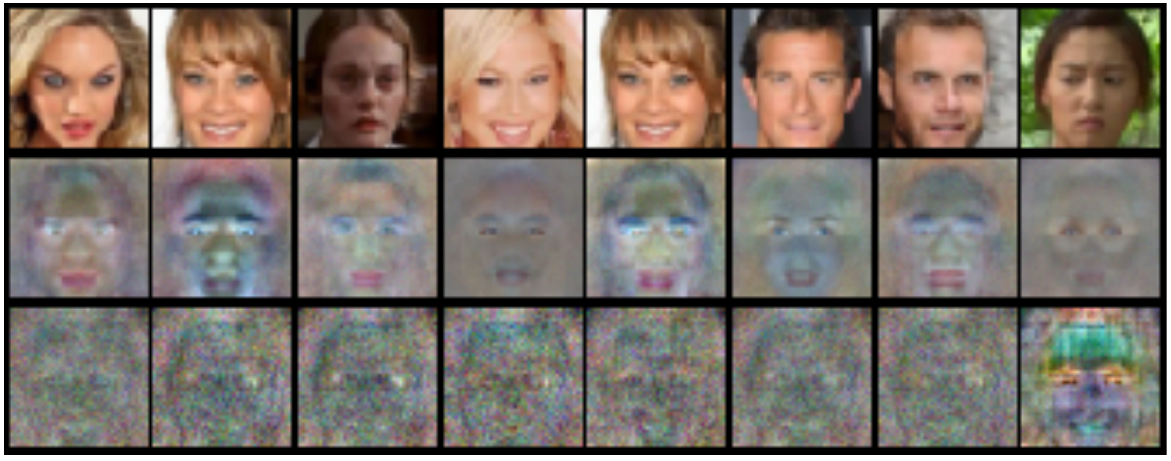}
    	\caption{Results of the model inversion attack. Top to bottom: real target images, reconstructions from the non-private model, reconstructions from the model trained by \texttt{FedGP}.}
	\label{fig:reconstruction}
\end{figure}

\begin{table}
	\caption{Face detection and recognition rates (pairs with distances below $0.99$) for images recovered by model inversion attack from the non-private baseline and the model trained by \texttt{FedGP}.}
	\label{tab:face_recognition}
	\centering
	\begin{tabular}{ c c c }
		\toprule
		{\bf } 				& {\bf Baseline} 	& {\bf FedGP} \\
		\midrule
		Detection 			& $25.5\%$ 			& $1.2\%$ 	\\
		Recognition 		& $2.8\%$ 			& $0.1\%$ 	\\
		\bottomrule
	\end{tabular}
\end{table}

On top of estimating expected privacy loss bounds, we test \texttt{FedGP}'s resistance to the \emph{model inversion attack}~\cite{fredrikson2015model}. More specifically, we run the attack on two student models: trained on original data samples and on artificial samples correspondingly. Note that we also experimented with another well-known attack on machine learning models, the membership inference~\cite{shokri2017membership}. However, we did not include it in the final evaluation, because of the poor attacker's performance in our setting (nearly random guess accuracy for given datasets and models even on the non-private baseline). Moreover, we only consider passive adversaries and we leave evaluation with active adversaries, e.g.~\cite{hitaj2017deep}, for future work.

In order to run the attack, we train a student model (a simple multi-layer perceptron with two hidden layers of 1000 and 300 neurons) in two settings: the real data and the artificial data generated by the federated GAN. As facial recognition is a more privacy-sensitive application, and provides a better visualisation of the attack, we pick the CelebA attribute prediction task to run this experiment.

We analyse real and reconstructed image pairs using OpenFace~\cite{amos2016openface} (see Table~\ref{tab:face_recognition}). It confirms our theory that artificial samples would shield real data in case of the downstream model attack. In the images reconstructed from a non-private model, faces were detected $25.5\%$ of the time and recognised in $2.8\%$ of cases. For our method, detection succeeded only in $1.2\%$ of faces and the recognition rate was $0.1\%$, well within the state-of-the-art error margin for face recognition.

Figure~\ref{fig:reconstruction} shows results of the model inversion attack. The top row presents the real target images. The following rows depict reconstructed images from the non-private model and the model trained on the federated GAN samples. One can observe a clear information loss in reconstructed images going from the non-private to the \texttt{FedGP}-trained model. Despite failing to conceal general shapes in training images (i.e. faces), our method seems to achieve a trade-off, hiding most of the specific features, while the non-private model reveals important facial features, such as skin and hair colour, expression, etc. The obtained reconstructions are either very noisy or converge to some average feature-less faces.

\section{Conclusions}
\label{sec:conclusion}
We study the intersection of federated learning and private data release using GANs. Combined these methods enable important advantages and applications for both fields, such as higher flexibility, reduced trust and expertise requirements on users, hierarchical data pooling, and data trading. 

The choice of GANs as a generative model ensures scalability and makes the technique suitable for real-world data with complex structure. In our experiments, we show that student models trained on artificial data can achieve high accuracy on classification tasks. Moreover, models can also be validated on artificial data. Importantly, unlike many prior approaches, our method does not assume access to similar publicly available data. 

We estimate and bound the expected privacy loss of an average client by using differential average-case privacy thus enhancing privacy of traditional federated learning. We find that, in most scenarios, the presence or absence of a single data point would not change the outcome probabilities by more than $1\%$ on average. Additionally, we evaluate the provided protection by running the model inversion attack and showing that training with the federated GAN reduces information leakage (e.g. face detection in recovered images drops from $25.5\%$ to $1.2\%$).

Considering the importance of the privacy research, the lack of good solutions for private data publishing, and the rising popularity of federated learning, there is a lot of potential for future work. In particular, a major direction of advancing current research would be achieving differential privacy guarantees for generative models while still preserving high utility of generated data. A step in another direction would be to improve our empirical privacy concept, e.g. by bounding maximum privacy loss rather than average, or finding a more principled way of sampling from outcome distributions.

\bibliographystyle{named}
\bibliography{fml_2019}

\end{document}